\documentclass[runningheads]{llncs}
\usepackage{esvect}
\usepackage[T1]{fontenc}
\usepackage{graphicx}
\usepackage{bbding} 
\usepackage{amsmath}
\usepackage{amssymb}
\usepackage{booktabs}
\usepackage{xcolor}
\newcommand{\xmark}{$\times$}
\usepackage{times}
\usepackage{soul}

\usepackage{multirow}
\usepackage{array}
\newcommand{\PreserveBackslash}[1]{\let\temp=\\#1\let\\=\temp}
\newcolumntype{C}[1]{>{\PreserveBackslash\centering}p{#1}}
\newcolumntype{R}[1]{>{\PreserveBackslash\raggedleft}p{#1}}
\newcolumntype{L}[1]{>{\PreserveBackslash\raggedright}p{#1}}
\usepackage{tikz}
\def\checkmark{\tikz\fill[scale=0.4](0,.35) -- (.25,0) -- (1,.7) -- (.25,.15) -- cycle;} 
\usepackage{float}
\usepackage{mwe}%for subfigure
\newcommand{\archiName}{DExNet}
\graphicspath{{Figures/}}
\newcommand{\etal}{\textit{et al}.}
\usepackage{orcidlink}

\begin{document}
\title{\archiName: Combining Observations of Domain Adapted Critics for Leaf Disease Classification with Limited Data}
\titlerunning{\archiName: Domain Adapted Critics for Leaf Disease Classification}

% \title{Contribution Title\thanks{Supported by organization x.}}
\author{Sabbir Ahmed\inst{1}\Envelope~\orcidlink{0000-0001-5928-4886} \and
Md. Bakhtiar Hasan\inst{1}\orcidlink{0000-0001-8093-5006} \and
Tasnim Ahmed\inst{1,2}\orcidlink{0000-0002-0799-1180} \and \\
Md. Hasanul Kabir\inst{1} \orcidlink{0000-0002-6853-8785}}
\authorrunning{S. Ahmed et al.}
\institute{
Department of CSE, Islamic University of Technology, Bangladesh\\
\email{\{sabbirahmed, bakhtiarhasan, tasnimahmed, hasanul\}@iut-dhaka.edu}
\and
School of Computing, Queen's University, Ontario, Canada\\
\email{tasnim.ahmed@queensu.ca}
}
% \\
% \email{sabbirahmed@iut-dhaka.edu}\\
% \email{sabbirahmed@iut-dhaka.edu}
%
\maketitle              % typeset the header of the contribution
\vspace{-5mm}
\begin{abstract}
% The abstract should briefly summarize the contents of the paper in
While deep learning-based architectures have been widely used for correctly detecting and classifying plant diseases, they require large-scale datasets to learn generalized features and achieve state-of-the-art performance. This poses a challenge for such models to obtain satisfactory performance in classifying leaf diseases with limited samples. This work proposes a few-shot learning framework, Domain-adapted Expert Network (\archiName), for plant disease classification that compensates for the lack of sufficient training data by combining observations of a number of expert critics. It starts with extracting the feature embeddings as `observations' from nine `critics' that are state-of-the-art pre-trained CNN-based architectures. These critics are `domain adapted' using a publicly available leaf disease dataset having no overlapping classes with the specific downstream task of interest. The observations are then passed to the `Feature Fusion Block' and finally to a classifier network consisting of Bi-LSTM layers. The proposed pipeline is evaluated on the 10 classes of tomato leaf images from the PlantVillage dataset, achieving promising accuracies of $89.06\%$, $92.46\%$, and $94.07\%$, respectively, for $5$-shot, $10$-shot, and $15$-shot classification. Furthermore, an accuracy of $98.09\pm0.7\%$ has been achieved in 80-shot classification, which is only $1.2\%$ less than state-of-the-art, allowing a $94.5\%$ reduction in the training data requirement. The proposed pipeline also outperforms existing works on leaf disease classification with limited data in both laboratory and real-life conditions in single-domain, mixed-domain, and cross-domain scenarios. 
%(The codes will be made publicly available upon the acceptance of the manuscript)
 %The idea is very simple- since we don't have enough training data, we have taken the observations of a number of expert critics, combined them, and produced a highly meaningful one. 

\end{abstract}
\vspace{-7mm}
\section{Introduction}
\vspace{-1mm}
\label{sec:intro}

%\subsection{Motivation and Scope}

%The advent of smart systems in the Agricultural domain has been a demand of time to achieve the expected harvest since the overall production is on the decline due to the crops being prone to various diseases \cite{panno2021review}. Traditional disease detection approaches require manual inspection of diseased leaves through visual cues or chemical analysis of infected areas, which can be susceptible to low detection efficiency and poor reliability due to human error. Adding to the problem, the lack of professional knowledge of the farmers and the unavailability of agricultural experts who can detect the diseases also hamper the overall harvest production. Negligence in this regard poses a significant threat to food security worldwide while causing great losses for the stakeholders. Early detection and classification of diseases implemented using tools and technologies available to the farmers can go a long way to alleviate all the issues discussed \cite{li2021plant}.
The decline in agricultural production due to diseases is a major threat to food security worldwide, causing severe losses for all stakeholders \cite{panno2021review}. Traditional disease detection methods are limited due to their low detection efficiency and poor reliability \cite{buja2021advances}. This is compounded by the lack of professional knowledge of farmers and the scarcity of agricultural experts in remote locations. As a result, the need for developing smart systems that can enable early detection and classification of diseases using tools and technologies available to the farmers can be pivotal \cite{li2021plant}.
% Several solutions have been proposed using the traditional machine learning approaches for plant disease classification \cite{liakos2018machine}. However, these works depend on handcrafted feature extraction techniques, which have failed to generalize on larger datasets. The emergence of deep learning-based methods in the agricultural domain has opened a new door for researchers with outstanding generalization capability removing the dependencies on extreme feature engineering \cite{kamilaris2018deep}. Recently, Convolutional Neural Network (CNN) has become a powerful tool for any classification task as it automatically extracts important features from images without human supervision and has been adapted to different leaf disease classification problems. Moreover, the recent variations of CNN architectures such as 
% AlexNet \cite{krizhevsky2012imagenet}, DenseNets \cite{huang2017densely}, EfficientNets \cite{tan2019efficientnet}, GoogLeNet \cite{szegedy2015going}, MobileNets \cite{howard2017mobilenets,sandler2018mobilenetv2}, NASNets \cite{zoph2018learning}, Residual Networks (ResNets) \cite{he2016deep}, SqueezeNet \cite{iandola2016squeezenet}, Visual Geometric Group (VGG) Networks \cite{simonyan2015very}, etc, have enabled the machines to understand complex patterns enabling even better performance than humans in many classification problems. 

Recent advancements in Deep Learning (DL) approaches using Convolutional Neural Networks (CNN) have become a powerful tool for several downstream classification tasks \cite{alanezi2022livestock,yasmeen2021csvcNet,rahman2022twoDecades,raiyan2025hasper,khan2022rethinking}. In particular, the utilization of transfer learning techniques using pre-trained architectures has provided numerous solutions in plant disease detection \cite{rafi2025mangoLeafVit,ashmafee2023apple,morshed2022Fruit,maeda2020comparison,abbas2021tomato}. However, most of these solutions propose deep and complex networks focusing on increasing detection accuracy, requiring a huge number of training images. Moreover, the challenge is exacerbated by the absence of substantial public datasets for categorizing rare plant diseases \cite{xu2023}. Thus, developing effective algorithms that can accurately detect diseases with limited data is crucial to improve the situation. One promising approach is Few-Shot Learning (FSL), which aims to enable models to generalize effectively from only a small number of labeled examples \cite{sun2024fslLeafSurvey,mehedi2024fslBHDR}. Although some FSL-based solutions for classifying leaf diseases have been suggested in the literature \cite{0018argueso2020fewshot,0510wang2021IMAL,0059yangLi2021metaLearning,0501Nuthalapati2021multi}, most of them have not demonstrated satisfactory performance and have proven ineffective in applications outside their original domain.

To this end, we utilize nine different state-of-the-art CNN-based feature extractors that are adapted to extract generalized features in the agriculture domain. These extracted features are then combined and passed through a classification block for effective disease prediction. %We followed a very simple intuition- as we don't have enough samples per class, we have to take multiple observations of the samples from a number of expert critics, with the intuition that each critic brings some unique description. If we can carefully combine those observations, we can come up with a better feature that leads to improved accuracy even with very limited samples per class.
Given the limited number of samples per class, we adopt an ensemble strategy leveraging multiple CNN-based feature extractors to obtain diverse and complementary representations of the input data. Each model contributes distinct feature perspectives, and their integration enhances the overall discriminative capacity of the system. This approach enables improved prediction performance despite the scarcity of labeled samples.
Our contributions can be summarized as follows:
\vspace{-0.2cm}

\begin{enumerate}
    \item We introduced domain-specific knowledge into pre-trained Deep CNN-based feature extractors in order to adapt them to the agriculture domain to compensate for target domain data scarcity.
    \item We incorporated a feature fusion block to combine
    %concatenate 
    extracted features from the critics to provide multi-faceted observation from limited samples.
    \item We utilized a Bi-LSTM architecture to capture the dependencies among the combined features to effectively classify leaf diseases.
    \item We outperformed the existing state-of-the-art FSL-based implementations in both laboratory and real-life conditions in single, multi, and cross-domain tasks.
\end{enumerate}

\vspace{-2mm}
\section{Related Work}
\vspace{-2mm}

The progressive integration of Deep Learning in agriculture requires a large amount of data to achieve reliable generalization, high classification accuracy, and resilience to variability in environmental conditions, crop types, and disease manifestations \cite{ahmed2024exeNet,herok2023cotton}. However, collecting and annotating such large-scale agricultural datasets is often labor-intensive, time-consuming, and infeasible for many real-world applications, particularly in regions with limited technical resources \cite{rafi2023pest}. However, few-shot learning (FSL) algorithms have shown promise in addressing this limitation by mimicking the human capacity to learn quickly with just a few labeled examples \cite{zhao2025fsl}. Different branches of FSL algorithms include metric-learning-based methods, parameter optimization, and data augmentation \cite{ahmed2022classification}.

In few-shot leaf image classification, 
a small labeled support set provides reference samples for each disease class, while an unlabeled query set is classified by comparing its images to the support set using distance metrics that are critical for accurate prediction under limited data scenarios\cite{UskanerHepsağ2024}. Metric-learning-based approaches utilize distance functions to measure the similarity and dissimilarity between leaf samples in the support set and the query set \cite{zabihzadeh2024zs}. The model produces an embedding vector from the input image, which is compared with the embeddings of other classes to predict the disease. 
Metric-based FSL approaches, such as Siamese Networks \cite{0101koch2015siamese}, Prototypical Networks \cite{1004snell2017prototypical}, MatchingNet \cite{1012vinyals2016matching}, etc., have shown superior performance over conventional CNN-based methods when dealing with limited data. Argüeso \etal \cite{0018argueso2020fewshot} employed an InceptionV3 backbone \cite{7780677} trained on 32 leaf disease classes to extract features for a Siamese Network using triplet loss, achieving competitive results with up to 90\% less training data. Similarly, Wang \etal \cite{0050wang2019plant} utilized a Siamese architecture with shared weights to extract features from image pairs, learning a similarity metric optimized via a contrastive loss, followed by K-NN classification. However, the high inter-class variation in their dataset reduced task complexity. To address more challenging scenarios, Egusquiza \etal \cite{0509itziar2022analysis} curated a real-world dataset of 5 crops and 17 diseases captured under varied and complex conditions, demonstrating that metric-based models outperform traditional CNNs even with fewer than 200 samples per class. Li \etal \cite{0059yangLi2021metaLearning} proposed a task-adaptive meta-learning framework for crop and pest classification, evaluating it under single-domain, mixed-domain, and cross-domain settings. Although their dataset aggregated samples from multiple public sources, its relatively low intra-class similarity simplified classification; nonetheless, performance degraded notably in cross-domain scenarios.

Parameter optimization-based techniques in FSL entail learning how to tune the parameters of a DL network using optimizers. Finn \etal \cite{1011chelsea2017model} proposed Model Agnostic Meta-learning (MAML), which solves few-shot tasks by learning an initialization parameter $\theta$. Once learned, new tasks can be solved with a few gradient updates. MAML is independent of the meta-learner algorithm used, making it ideal for machine learning algorithms that require quick adaptation. 
Other improvements on MAML include TAML \cite{1028jamal2019taskAgnostic}, Reptile \cite{1029nichol2018reptile}, etc. 
In this regard, Wang \etal \cite{0510wang2021IMAL} exploited the MAML approaches with soft-center loss function and PReLU activation function for leaf disease classification. The proposed method outperformed other FSL approaches and achieved 63.8\%, 91.35\%, and 96.0\% accuracy for 1-shot, 15-shot, and 80-shot classification, respectively. However, the approach requires a large number of tasks and consumes high computational resources, and further optimization is needed to increase speed and reduce computational overhead.

Expanding the number of samples via data augmentation can solve the lack of labeled samples to some extent. Hu \etal \cite{0017hu2019aLowShot} proposed an FSL method for tea leaf disease identification that uses SVM for feature extraction and C-DCGAN for data augmentation. The augmented samples were used to train the VGG16 architecture \cite{simonyan2014very}, achieving an accuracy of up to 90\% on a self-curated dataset of three tea leaf diseases.
In another work, Nesteruk \etal \cite{0070sergey2021image} claimed that the use of such augmentation techniques can improve the performance of other downstream tasks involving leaf diseases, such as object detection, segmentation, contouring, denoising, etc. %However, the authenticity of the augmented samples can be questionable.
However, the authenticity of the augmented samples in reflecting real-world data distributions remains uncertain.

Recent studies have demonstrated that transfer learning with pre-trained architectures can effectively generate generalizable feature representations for few-shot learning tasks. Chen \etal \cite{0103chen2019aCloseLook} proposed a simple transfer learning-based approach using pre-trained CNN and linear classifiers that outperforms traditional FSL methods. Kolesnikov \etal \cite{1017Kolesnikov2020bigTransfer} suggested that a high-quality feature extractor is the best way to solve FSL tasks and proposed an approach to train a large model on a large dataset to learn the feature space.  Yonglong \etal \cite{0104yonglong2020rethingkingFewShot} also demonstrated that a simple transfer-based approach can outperform advanced meta-learners. Dvornik \etal \cite{1018dvornik2020selectingRelevant} introduced diversity in feature extraction by using a set of feature extractors trained on different datasets. They used ensembling to train multiple deep CNNs during the meta-training phase and encouraged diversity in the learned representations. 

In the domain of agriculture, Nuthalapati \etal \cite{0501Nuthalapati2021multi} proposed a method for FSL that utilizes a ResNet18-based feature extractor pre-trained on ImageNet. A transformer block is used to further enhance the feature embeddings of all support set samples. 
%During the classification phase, Mahalanobis distance is used to calculate the similarity between transformed embeddings and the embedding of the image to be classified. 
During classification, similarity between the query image embedding and the class prototypes is computed using the Mahalanobis distance. The effectiveness of the model was shown using mixed-domain and cross-domain experiments using the Plant and Pest \cite{0059yangLi2021metaLearning} and the Tomato leaf subset of the Plantvillage dataset \cite{hughes2015open}; where the proposed model achieved up to 14\% and 24\% performance gain compared to the earlier works. Moreover, to validate the performance of the model in real-world images, the authors curated the `Plants in Wild' dataset and provided a baseline.

% Despite recent works using transfer learning-based approaches in FSL domain, these methods have yet to achieve high performance, and their accuracy degrades further when exposed to samples beyond laboratory conditions. Our work leverages a set of pre-trained feature extractors and adapts them with domain-specific knowledge, followed by feature fusion and classifier blocks, to enable robust and generalizable leaf disease classification under limited data scenarios, ensuring state-of-the-art performance.
While transfer learning-based approaches have gained traction in the FSL domain, their performance remains constrained, particularly in real-world scenarios, due to restricted feature diversity from relying on a single pre-trained model. To address these challenges, we integrate multiple pre-trained feature extractors adapted with domain-specific knowledge, followed by feature fusion and classification modules. This enables robust and generalizable leaf disease classification under limited data conditions, achieving state-of-the-art performance.

\vspace{-2mm}
\section{Task Formulation}
\vspace{-2mm}

Few-shot learning (FSL) experiments have two phases: meta-training and meta-testing. The datasets used in these two phases have non-overlapping classes. Each of them is divided into Support sets ($S$) and Query sets ($Q$), similar to the train-test splits of the traditional DL experiments. 

The meta-training set is used to train feature extractors that learn to generate discriminative embeddings from limited samples, with no restriction on the number of classes or examples, provided there is no overlap with the meta-testing set. The meta-testing set contains $N$ classes, each represented using $k$ labeled examples extracted from the Support set (i.e., $k$-shot). After training the classifier using these few samples, performance is evaluated on $Q$ unseen examples per class drawn from the Query set. To ensure statistical robustness, models are evaluated across multiple tasks, each defined by randomly sampled Support and Query sets.

% The meta-training dataset is used to train the feature extractors to learn to produce good feature embeddings from limited samples. The number of classes and samples in this portion is not limited as long as it does not overlap with the meta-testing portion. In contrast, the meta-testing dataset consists of $N$ classes. The Support set consists of very `few' samples, from which $k$ samples are taken to train or fine-tune the classifier, known as `$k$-shots'. After training, $Q$ unseen samples are taken from the Query set, and the overall performance is evaluated. The models are usually tested for multiple such `tasks' with randomly chosen Support and Query samples for statistical significance.

To comprehensively assess the FSL pipeline, single-domain, mixed-domain, and cross-domain experiments are conducted. In single-domain settings, the meta-training and meta-testing datasets originate from similar domains with disjoint classes. Cross-domain experiments involve meta-training and meta-testing datasets drawn from distinct domains, presenting increased difficulty due to significant domain distribution discrepancies. Mixed-domain experiments include classes from multiple domains in both meta-training and meta-testing phases.
% To further evaluate the effectiveness of the FSL pipeline, single-domain, mixed-domain, and cross-domain experiments are conducted. In the single-domain experiments, the meta-training and meta-testing \COR{datasets} come from similar domains without any class overlaps. On the contrary, for the cross-domain experiments, the datasets of the meta-training and meta-testing phases come from two different domains. This task is even more challenging since it is very difficult to produce feature distribution by tackling the high amount of dissimilarity of the domains of interest. In mixed-domain FSL, both the meta-training and meta-testing sets contain classes from multiple domains.

For the single-domain experiments, we have used the PlantVillage Dataset \cite{hughes2015open}, the largest open-access repository of expertly curated samples for leaf disease classification tasks, consisting of 38 classes. %belonging to 14 crops. 
Out of these classes, we have picked the 10 classes of tomato leaf images for the meta-testing phase and used the rest of the 28 classes for domain adaptation. In the meta-training phase, the 28 classes have been used to fine-tune the feature extractors to enhance their capabilities for leaf disease identification. On the other hand, the 10 classes of tomato leaf samples have been divided into Support and Query sets with an $80:20$ ratio following the work of Ahmed \etal \cite{9810234} Hence, out of the 18160 samples of tomato leaf images, 3629 samples have been considered as the available samples as the Query set, and the remaining images have been used as the Support set. For each task, $k$ Support samples and $Q$ query samples per class are randomly picked from the available images. To remove biases, this process has been repeated 100 tasks, and the average performance has been reported. 

For the Mixed-domain and Cross-domain experiments, we have utilized the `Plants \& Pest' dataset proposed by Li \etal \cite{0059yangLi2021metaLearning}. The dataset consists of 20 classes in total, out of which 10 classes belong to different kinds of plants, such as apple, blueberry, cherry, grape, etc. The other 10 classes contain images of different kinds of pests, such as cydia-pomonella, fruit-fly, gryllotalpa, snail, etc. Each of the classes contains 300 images. We have followed the recommended splits by the original authors for the mixed and cross-domain experiments \cite{0059yangLi2021metaLearning}. In the mixed-domain tasks, we have taken 5 plants and 5 pest classes for meta-training and the rest of the classes for meta-testing. For the Cross-domain experiments, we have conducted two kinds of experiments. In the first one, the meta-training set consists of the 10 pest classes, and in another cross-domain scenario, we have considered all 10 plant classes for meta-training and the remaining classes for meta-testing. 

Furthermore, to evaluate the efficacy of the proposed pipeline in real-world field images, we conducted experiments on two publicly available datasets, `Potato Leaf Disease Classification'\footnote{https://github.com/rizqiamaliatuss/PotatoLeafDiseaseClassification}\cite{potato5cls} and `Cotton Leaf Disease'\footnote{https://www.kaggle.com/datasets/seroshkarim/cotton-leaf-disease-dataset}\cite{cotton4cls}, consisting of diversified samples collected from the field. All these experiments ensure that our proposed pipeline has been validated under diverse and challenging conditions, demonstrating its robustness, adaptability, and generalization capabilities across single-domain, mixed-domain, cross-domain, and real-world field scenarios.

\vspace{-2mm}
\section{\archiName}
\vspace{-1mm}
% Deep Learning-based solutions have achieved state-of-the-art performance in different leaf disease classification tasks, but with the requirement of huge labeled data. However, this high performance poses the requirement of a huge amount of training data, which is a very difficult constraint to satisfy. 
% \archiName\ alleviates this limitation by producing high performance even with very limited samples. It consists of two major components, the `Domain-Adapted Feature Extraction Block' and the `Feature Fusion Network' (FFN). 
% % Domain-adapted feature extraction block
% At first, the input is passed to the feature extractor block, which consists of nine state-of-the-art pre-trained Deep CNN architectures. Each of the CNN components is `Domain-adapted' for improved feature extraction. The input is passed through each of the CNN blocks and a feature vector is extracted from each of them. In this way, for each input, the feature extraction block produces nine `observations' describing the different characteristics. 
% % Featuer Fusion Network
% The features are then passed to the Feature Fusion network, where all nine observations are concatenated to produce a combined feature vector and then passed through a Classifier block for the final prediction. We explored a variety of choices for the classifier block and found the best result by implementing a Bi-LSTM layer with 1024 nodes. 
% Finally, the classifier block is densely connected with a softmax layer which provides the final prediction. 
The Domain-adapted Expert Network (\archiName) pipeline addresses the challenge of limited labeled samples per class by aggregating insights from multiple specialized expert models, or ``critics.'' These critics are exposed to a broad distribution of leaf images during domain adaptation, enabling them to learn generalizable features of diseased leaves. Although they are not trained on the target classes, their learned representations remain descriptive and informative, allowing the model to perform effectively even with very few target samples.
% The Domain-adapted Expert Network (\archiName)\ pipeline follows a very simple intuition- as we do not have enough samples per class, we compensate by taking `observations' from a number of expert `critics' and by combining them, we can get an overall understanding of the image. To enhance the descriptive ability of the critics, we show them many leaf images in the domain adaptation process. Hence, although our critics have never seen the samples of target classes, they know how to describe a diseased leaf in general, and so, they can provide quality observations even from `very few samples'. 

\begin{figure*}[tb]
    \centering
    \includegraphics[width=0.8\textwidth]{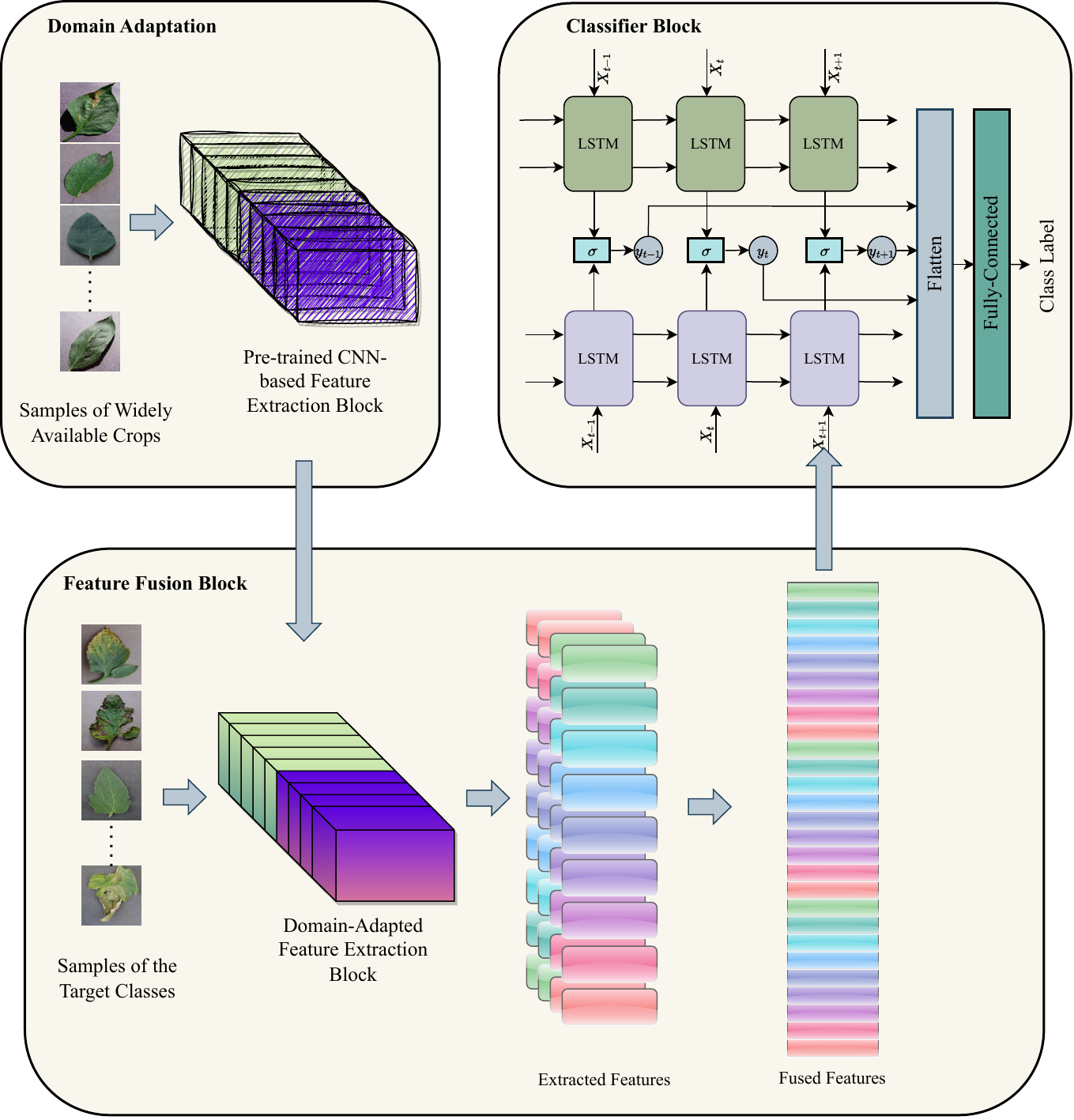}
    \caption{Overview of the Proposed Pipeline \archiName\ }
    \label{fig:fslPipeline}
    \vspace{-5mm}
\end{figure*}

\archiName\ comprises three main components: the Domain Adaptation Block (DAB), the Feature Fusion Block (FFB), and the Classifier Block (CB). In the DAB, we fine-tune nine pre-trained deep CNN models (referred to as \textit{critics}) using a diverse set of publicly available leaf images. This domain adaptation process enables the critics to specialize in extracting disease-relevant features, effectively transforming them into \textit{experts}. 
In the FFB, feature representations (termed \textit{observations}) are extracted by passing target-class samples through each of the domain-adapted experts. The resulting nine feature vectors are then concatenated to form a unified representation. Finally, the CB employs a Bi-LSTM layer with 1024 nodes followed by a fully-connected layer to perform the final classification. \figureautorefname~\ref{fig:fslPipeline} illustrates the proposed pipeline \archiName.

% \archiName\ consists of three components: Domain-Adaptation Block (DAB), Feature Fusion Block (FFB), and Classifier Block (CB). The DAB utilizes samples of widely available leaf samples to fine-tune nine pre-trained Deep CNN models that we term as `critics' for adaptation to agriculture domain-specific feature extraction. Hence, the `critics' become `experts' by going through the process of `Domain Adaptation'. 
% In FFB, the samples of the target classes are first passed through the domain-adapted CNN-based feature extraction block. Each CNN model (`experts') produces a `feature vector' that we term as `observations', and these nine observations are `concatenated' in the FFB to generate a combined feature vector. The Classifier block with a Bi-LSTM layer and 1024 nodes are used for the final prediction. A pictorial view of the proposed pipeline is shown in \figureautorefname~\ref{fig:fslPipeline}.

% In the proposed pipeline, the CNN-based feature extractor produces a concatenated feature vector for each input.  It is then passed to the classifier block to find deeper meaning and produce class predictions. To enhance the feature learning ability of the extractors, we have introduced the concept of domain adaptation. In the following discussions, we have explained each of these components and provided justification behind each design choice. 

\vspace{-1mm}
\subsection{CNN-Based Feature Extractor}
While conventional DL methods rely on large amounts of data to achieve satisfactory performance, such data requirements are impractical in FSL scenarios. To address this limitation, we combine the extracted features from several Deep CNN architectures, each with its unique method of feature extraction, to generate multiple observations. We hypothesize that showing a few samples to multiple critics and deciding the result based on their aggregated observations can provide a %comprehensive
better understanding of the samples. Here, the feature extractor is compared to a \textit{critic}, and the features produced are the \textit{observations} of the leaf images. This approach enables using spatial information in the image to generate feature vectors for training an FSL pipeline.

Our CNN-based Feature Extraction Block consists of state-of-the-art pre-trained residual and dense CNN architectures. The Residual Networks (ResNets) \cite{he2016deep} offer significant improvements over previous architectures to facilitate the training of many stacked layers by introducing residual connections between layers, allowing information to flow from shallow layers to deeper layers and gradients to flow back, skipping a few layers. We incorporated five pre-trained variants of ResNets in our domain-adapted expert feature extractor block, namely ResNet-18, ResNet-34, ResNet-50, ResNet-101, and ResNet-152. Expanding on this idea, Dense Convolutional Network (DenseNet) \cite{huang2017densely} connects all layers with matching feature-map sizes directly with each other. Here, each layer receives input from all the previous layers and sends its output to all the subsequent layers via concatenation. Creating a dense connection between all the layers ensures the maximum flow of information and gradient between layers. Four variants of the DenseNet architecture were chosen based on the number of layers and their computational complexity: DenseNet-121, DenseNet-161, DenseNet-169, and DenseNet-201.
%These models have been chosen considering their remarkable performance in large-scale image classification tasks \cite{0516Chowdhury2021ICCV}. 
Each of the models is pre-trained on the ILSVRC2012 dataset, and the final prediction layers were removed from the feature extractors to retain the feature embeddings.

\vspace{-2mm}
\subsection{Domain Adaptation}
\vspace{-1mm}
The feature extraction architectures are capable of extracting highly sophisticated features that can be utilized for any classification problem. When these networks are used for any classification task, a common practice is to fine-tune the architectures based on the available training images. However, in the FSL problems, it is not possible to fine-tune such deep models due to the limited number of samples in the Support set. An alternative solution to this problem is to provide knowledge of a similar domain so that, despite the model having never directly seen the target classes, it can have some idea about their general characteristics--- a process we term as \textit{Domain Adaptation}.

% domain adaptation kivabe korlam...
To apply domain adaptation in the proposed pipeline, we have trained the CNN-based architectures with a substantial amount of leaf images containing healthy and diseased samples. For this, we used the 28 classes of the PlantVillage dataset except for the samples of the tomato leaves. We hypothesize that since the models were already capable of extracting useful features, this knowledge base can be further enriched by fine-tuning them using a lot of leaf samples. After fine-tuning, even though the model has never seen any `tomato leaf samples', it knows what a healthy or diseased leaf looks like, what features to look for in a leaf image, and how to differentiate the classes of diseases, etc, turning it into an \textit{expert} model. Thus, when limited samples are provided, the expert models can extract even better features compared to their earlier states.

\vspace{-2mm}
\subsection{Feature Fusion Block (FFB)}
\vspace{-1mm}
% We employed nine state-of-the-art CNN-based models in our domain-adapted feature extraction block, yielding a total of nine feature vectors for each input. These features can also be thought of as the observations made by the experts (i.e., models). Since every critic expert has their unique capabilities, each observation provides a slightly different description of the input. Hence, we combine the features before passing them to the classifier with the expectation to find out even more meaningful features. We concatenated all the observations to produce a $ 13,984$-dimensional feature vector per sample. Since this approach preserves all the information in the individual feature vectors, it allows the model to learn complex interactions between the features of the same sample.
We leverage nine domain-adapted expert models for feature extraction, resulting in nine distinct feature vectors for each input image. These feature vectors (referred to as \textit{observations}) reflect the individual perspectives of the experts, with each model capturing unique characteristics of the input. To enhance the representational power, we concatenate all nine observations into a single $13,984$-dimensional feature vector per sample. This fusion approach preserves the distinct information captured by each expert, enabling the model to learn complex interactions among the features and facilitating the extraction of more informative and comprehensive representations.

%%%%%%%%%%%%% Classifier Network %%%%%%%%%%%
\vspace{-2mm}
\subsection{Classifier Block (CB)}
\vspace{-1mm}

%The CNN-based feature extractor block produces nine different feature observations using the pre-trained CNN models which are concatenated to a feature vector of $xxxx$ dimension. This vector is passed to the classifier network for the final prediction. 

%We have explored several choices of the Classifier networks. At first, we started with using Dense Block(s). \COR{(add text on what was our intuition and how it turned out.) } Since the features from different CNN models were concatenated on top of each other, it was necessary to take combinations of these ... 

%\COR{summarize the need of Bi-LSTM from link pasted in sheet}

The Classifier Block processes the concatenated feature vector from FFB using a Bi-LSTM layer with 1024 units \cite{graves2005framewise}, followed by a fully connected layer to produce the final classification output. The main purpose of this block is to effectively combine the diverse observations from the expert models and capture intricate relationships within the high-dimensional feature vector. The Bi-LSTM layer consists of two LSTM networks running in parallel, processing the inputs in opposite directions. This structure allows the model to capture both local dependencies between adjacent elements and the broader contextual information of the entire sequence, enhancing the model's ability to learn complex interactions. Upon empirical evaluation with varying dimensions (e.g., 256, 512, 1024, 2048), we found that 1024 units provided the best performance. 
\vspace{-2mm}
\section{Results \& Discussions}
\vspace{-1mm}
\subsection{Single vs Ensemble Feature Extractor}
The performance of each of the feature extractors was thoroughly analyzed for different choices of $k$ shots. Along with finding the performance of the individual networks, we also considered an ensemble of ResNets, an ensemble of DenseNets, and finally, the ensemble of all nine architectures. Since the focus was on finding the best combination of feature extractors, we only added a Dense layer of $1024$ hidden units after the flattened feature vector from the pretrained extractors, followed by a softmax layer. 
We experimented with four combinations of $k = 1, 5, 10,\& 15$. For the test samples, $Q = 50$ images per class were taken at random. To remove any bias, this random sampling was repeated for 100 separate trials, and the average value was reported. The findings of this experiment are shown in  \tableautorefname~\ref{tab:singleVsEnsembel}.
% we focused on finding the best combination of feature extractors, we used only a Dense layer of $1024$ hidden units followed by a softmax layer. 
\begin{table}[htb]
    \caption{Performance of Single vs Ensemble Feature Extractors}
    \vspace{-6mm}

    \label{tab:singleVsEnsembel}
    \begin{center}
    \begin{tabular}{L{0.25\textwidth} C{0.18\textwidth}         C{0.18\textwidth} C{0.18\textwidth} C{0.18\textwidth}}
        % \begin{tabular}{L{2cm} C{1cm} C{1cm} C{1cm} C{1cm}}
        \toprule
        \textbf{Feature Extractor} 
        % & \boldmath$K=1$ & \boldmath$K=5$ & \boldmath$K=10$ & \boldmath$K=15$\\
         & \textbf{\textit{k} = 1} & \textbf{\textit{k} = 5} 
            & \textbf{\textit{k} = 10} & \textbf{\textit{k} = 15}\\
        \midrule
        ResNet18 & $34.36\pm0.82$ & $55.42\pm0.77$ & $62.56\pm0.67$ & $66.76\pm0.62$\\
        ResNet34 & $29.65\pm0.91$ & $49.93\pm0.81$ & $57.93\pm0.56$ & $63.69\pm0.54$ \\ 
        ResNet50 & $30.62\pm0.88$ & $54.24\pm0.92$ & $61.79\pm0.65$ & $65.99\pm0.64$ \\ 
        ResNet101 & $29.64\pm0.93$ & $52.73\pm0.79$ & $61.61\pm0.74$ & $65.79\pm0.68$ \\ 
        ResNet152 & $29.75\pm0.84$ & $52.61\pm0.86$ & $61.39\pm0.74$ & $66.44\pm0.68$ \\ 
        \midrule
        DenseNet121 & $32.48\pm0.99$ & $54.93\pm0.89$ & $64.09\pm0.72$ & $68.21\pm0.56$ \\ 
        DenseNet161 & $32.29\pm1.07$ & $56.58\pm0.88$ & $65.59\pm0.71$ & $70.56\pm0.66$ \\ 
        DenseNet169 & $32.89\pm0.97$ & $55.39\pm0.78$ & $64.09\pm0.72$ & $69.39\pm0.7$ \\ 
        DenseNet201 & $33.51\pm0.94$ & $57.21\pm0.87$ & $65.71\pm0.86$ & $70.77\pm0.73$ \\ 
        \midrule
        Ensemble of ResNets & $36.34\pm1.01$ & $60.04\pm0.91$ & $68.74\pm0.71$ & $73.84\pm0.68$ \\ 
        \midrule
        Ensemble of DenseNets & $35.94\pm1.03$ & $61.47\pm0.96$ & $70.26\pm0.79$ & $75.42\pm0.65$\\
        \midrule
        Ensemble of Nine Extractors & \boldmath$37.89\pm0.99$ & \boldmath$62.07\pm0.98$ & \boldmath$72.33\pm0.81$ & \boldmath$76.93\pm0.66$ \\
        \bottomrule
    \end{tabular}
    \vspace{-5mm}
    \end{center}
\end{table}

The objective was to find the feature extractor that works best with the provided `few-shots'. Since the performance of the individual networks was close to each other, we hypothesized that ensembling them could provide better results because each of them has unique strengths in extracting certain types of features useful for detecting specific types of diseases. By ensembling similar types of networks, we achieved better performance than the individual ones. The best performance was achieved by combining all nine feature extractors.

\vspace{-1mm}
\subsection{Combination of the Extracted Features}

% Once the leaf images are given into the proposed pipeline, the feature extraction block provides nine unique observations. These features are then combined and sent to the classifier block to find out deeper meanings. To combine the features, we have attempted two approaches; which we have named, `Parallel Observation' and `Concatenated Observation'.
The features extracted from the domain-adapted feature extraction blocks are subsequently combined and passed to the classifier block for both training and evaluation phases. To identify an effective feature integration strategy, we consider two approaches: `Parallel Observation' and `Concatenated Observation'. In the Parallel Observation approach, each of the nine feature vectors, one from each feature extractor, is treated as an independent input to the classifier. This can be viewed as nine distinct observers offering complementary perspectives on the same image. In contrast, the Concatenated Observation approach merges all nine extracted features into a single high-dimensional representation by concatenation, resulting in a $13,985$-dimensional vector per sample. This unified representation enables the classifier to learn cross-feature dependencies and integrate multi-perspective cues into a single, discriminative embedding. In both cases, given the architectural heterogeneity of the feature extractors, each vector encodes unique complementary information, which enhances the classifier's ability to generalize by implicitly increasing the diversity of the input space.
% The features extracted by the domain-adapted feature extraction blocks are combined and passed to the classifier block for training and evaluation. To combine the features, we experimented with two approaches: `Parallel Observation' and `Concatenated Observation'. In the Parallel Observation approach, the nine observations, i.e, the nine different features extracted from each sample are passed to the classifier network separately. This can be interpreted as if nine observers are describing the image to the classifier. Since the inherent architecture of each of the feature extractors differs, each of the observations carries some `unique' information about the sample. It provides the model with an opportunity to create an effect of nine times more samples than what is provided. 
% On the other hand, the Concatenated Observation approach combines the nine features into one observation by concatenating with one another, resulting in a $13,984$ dimensional feature vector per sample. The objective of the classifier is to pick the most important information from the different observations of the feature extractors and combine them to produce a fine-tuned feature vector.

For each approach, two types of classifiers have been used: Dense Classifier with $1024$ nodes, and Bi-LSTM classifier with $1024$ units. We have also considered the impact of domain adaptation for each of the combinations. For this experiment, we have considered the values of $k$ as $1, 5, 10 \& 15$ and $Q = 50$ applied for 100 tasks. The findings are mentioned in \tableautorefname~\ref{tab:combineFeatures}.
For each combination, the domain-adapted feature extractors have provided $5-10\%$ performance improvement compared to the pre-trained ones. The Dense classifier is a better fit for the parallel observation approach, whereas the Bi-LSTM classifier is more suitable for the feature concatenation approach. Out of all the experiments, the Domain Adapted Concatenated feature extraction approach with Bi-LSTM classifier is the best performing one, having an accuracy as high as 90.97\% and 94.02\% respectively for 10-shot and 15-shot classification.

\begin{table*}[tb]
    \centering
    \caption{Performance of Various Combinations of Features in Different FSL Scenarios}
    \label{tab:combineFeatures}    
    % \begin{tabular}{L{0.15\textwidth} C{0.11\textwidth} C{0.1\textwidth}  C{0.16\textwidth} C{0.16\textwidth} C{0.16\textwidth} C{0.16\textwidth}}
    \begin{tabular}{L{0.15\textwidth} C{0.11\textwidth} C{0.1\textwidth}  C{0.11\textwidth} C{0.11\textwidth} C{0.11\textwidth} C{0.11\textwidth}}
        \toprule
        \textbf{Observation Type} & \textbf{Classifier} & \textbf{Domain Adopt} & \boldmath$k=1$ & \boldmath$k=5$ & \boldmath$k=10$ & \boldmath$k=15$\\
        \midrule
        Parallel & Dense & \xmark 
        & $40.2\pm1.46$ & $72.95\pm0.97$ 
        & $82.6\pm0.97$ & $85.71\pm0.78$\\
        Parallel & Dense & \checkmark 
        & $50.38\pm1.68$ & $83.78\pm1.16$ 
        & $89.07\pm0.91$ & $92.79\pm0.69$\\
        \midrule
        Parallel & Bi-LSTM & \xmark 
        & $39.84\pm1.65$ & $48.27\pm2.20$
        & $52.66\pm1.75$ & $63.42\pm1.78$\\
        Parallel & Bi-LSTM & \checkmark
        & $53.4\pm1.74$ & $74.75\pm1.68$
        & $79.3\pm1.90$ & $81.99\pm1.81$\\
        \midrule
        Concatenated & Dense & \xmark 
        & $37.89\pm0.99$ & $62.07\pm0.98$ 
        & $72.33\pm0.81$ & $76.93\pm0.66$\\
        Concatenated & Dense & \checkmark
        & $46.56\pm1.07$ & $72.17\pm1.04$
        & $81.18\pm0.68$ & $85.12\pm1.02$\\
        \midrule
        Concatenated & Bi-LSTM & \xmark
        & $42.53\pm1.25$ & $81.9\pm1.37$
        & $87.80\pm1.23$ & $89.06\pm1.22$\\
        Concatenated & Bi-LSTM & \checkmark
        & \boldmath$51.36\pm1.54$ & \boldmath$89.19\pm1.20$
        & \boldmath$90.75\pm1.09$ & \boldmath$94.02\pm0.86$\\
        \bottomrule
    \end{tabular}
\end{table*}

%%%%%%%% Classifier network %%%%%%%%%%
% \vspace{-2mm}
\subsection{Selection of Classifier Network}
% \vspace{-1mm}
The performance of Dense, LSTM, GRU, Bi-GRU, and Bi-LSTM classifiers was evaluated to determine the most effective classifier network. Each of the classifiers consisted of 1024 nodes followed by a softmax layer of 10 nodes, representing the 10 classes of the target samples. Experiments have been conducted on the base feature extractors without applying domain adaptation, with $k=1, 5, 10, and~15$. For this analysis, we considered the entire test set, comprising a total of 3629 images ($Q=Full$). The findings have been reported in \tableautorefname~\ref{tab:choosingClassifier}. 

\begin{table}[htb]
    \centering
    \caption{Performance of Different Classifier Networks}
    \label{tab:choosingClassifier}
    \begin{tabular}{L{0.15\textwidth} C{0.18\textwidth} C{0.18\textwidth} C{0.18\textwidth} C{0.18\textwidth}}
    % \begin{tabular}{L{1.4cm} C{1cm} C{1cm} C{1cm} C{1cm}}
          \toprule
          \textbf{Classifier} 
          % & \boldmath$K=1$ & \boldmath$k=5$
          % & \boldmath$K=10$ & \boldmath$K=15$\\
          & \textbf{\textit{k} = 1} & \textbf{\textit{k} = 10} 
            & \textbf{\textit{k} = 15} & \textbf{\textit{k} = 80}\\
          \midrule
          Dense Layer
          & $37.73\pm0.49$ & $61.08\pm0.88$
          & $73.54\pm0.72$ & $75.79\pm0.81$\\
         % Dense 
         %  & $37.89\pm0.99$ & $62.07\pm0.98$
         %  & $72.33\pm0.81$ & $76.93\pm0.66$\\
          LSTM
          & $43.24\pm1.72$ & $69.16\pm1.75$
          & $74.15\pm1.52$ & $72.84\pm1.02$\\
          GRU
          & $40.21\pm1.55$ & $68.15\pm1.54$
          & $78.97\pm1.54$ & $81.08\pm1.56$\\
          Bi-GRU
          & $43.03\pm0.51$ & $72.69\pm1.28$
          & $85.53\pm1.26$ & $90.94\pm1.03$\\
          Bi-LSTM
          & \boldmath$48.02\pm1.63$ & \boldmath$84.36\pm1.62$
          & \boldmath$88.97\pm1.80$ & \boldmath$91.22\pm1.67$\\
          \midrule
    \end{tabular}
    \vspace{-0.4cm}
\end{table}

% \COR{(why is dense underperforming? intuition behind using LSTM? )}

% RNN-based classifiers are well-suited for tasks involving structured or sequential inputs, as they can model long-range dependencies across feature elements. In contrast, dense classifiers treat features independently and are limited in capturing inter-feature relationships. In our setup, the concatenated features encode complementary information, and modeling their interaction is crucial for accurate classification. RNNs, particularly LSTM-based architectures, are better equipped to capture these dependencies and thus outperform dense counterparts.
RNN-based classifiers are typically better suited for tasks that involve sequential data, as they can capture long-term dependencies and relationships between elements in a sequence. However, dense classifiers provide higher importance to each feature separately, but they are unable to capture spatial relationships among them. In our case, capturing the spatial relationships among the concatenated features can provide a better understanding of the leaf samples from which they were extracted. Since RNN-based architectures are better suited for capturing important information from the observations of the expert critics, their performances are significantly better than Dense classifiers.

% Among the RNN-based architectures, LSTMs generally perform better than GRU due to their ability to store more long-term dependencies while also capturing more complex representations of the feature space. Learning to extract useful features from a limited number of leaf samples requires a higher level of expressivity, which can be attained by LSTM architectures, as evident from the results. 

Among RNN variants, Bi-LSTM offers additional benefits by processing the input in both forward and backward directions, which is particularly useful in our context where the disease class may depend on interactions across the entire feature sequence. Furthermore, Bi-LSTM demonstrates improved performance when handling long input vectors, as is the case with our high-dimensional concatenated features, outperforming both unidirectional LSTM and GRU classifiers.
% Compared to LSTM, Bi-LSTM layers can capture dependencies in both the forward and backward directions of the feature sequence, which has been especially useful in our task where the disease class of a certain leaf can depend on the observations of all the critics. Bi-LSTM is better at capturing the relationship between distant elements of the feature vector. At the same time, considering the large size of the concatenated feature vector, Bi-LSTMs have been shown to outperform LSTM and other RNN-based classifiers while dealing with long input sequences.
We also explored attention-based classifier variants but observed degraded performance, likely due to the limited size of the training data. As a result, we adopted Bi-LSTM as the classifier of choice.

\vspace{-2mm}
\subsection{Impact of Domain Adaptation}

The baseline CNN-based feature extractors are pre-trained on large-scale ImageNet datasets with the capability of extracting useful features from any input samples. However, to make them even more specialized in finding appropriate leaf features, we applied domain adaptation, which significantly improved the overall performance of the pipeline. We have used the Bi-LSTM classifier since it is most useful on the feature concatenation approach. The experimental findings are available in \tableautorefname~\ref{tab:domainAdaptation}.

\begin{table}[htb]
    \centering
    \caption{Impact of Domain Adaptation}
    \label{tab:domainAdaptation}
    \begin{tabular}{L{0.15\textwidth} C{0.2\textwidth} C{0.2\textwidth}}
        \toprule
        \textbf{k-shot} & \textbf{Domain Adaptation} 
        & \boldmath$Q = Full$\\
        \midrule
        \multirow{2}{0.12\textwidth}{$k=1$}  
        & \xmark & $48.02\pm1.63$\\
        & \checkmark & $56.19\pm1.72$\\
        \midrule
        \multirow{2}{0.12\textwidth}{$k=5$}  
        & \xmark & $84.36\pm1.62$\\
        & \checkmark & $89.06\pm1.30$\\
        \midrule
        \multirow{2}{0.12\textwidth}{$k=10$}  
        & \xmark & $88.97\pm1.80$\\
        & \checkmark & $92.46\pm1.75$\\
        \midrule
        \multirow{2}{0.12\textwidth}{$k=15$}  
        & \xmark & $91.22\pm1.67$\\
        & \checkmark & $94.07\pm1.40$\\
        \midrule
        \multirow{2}{0.12\textwidth}{$k=20$}  
        & \xmark & $92.66\pm1.87$\\
        & \checkmark & $95.68\pm1.20$\\
        \midrule
        \multirow{2}{0.12\textwidth}{$k=40$}  
        & \xmark & $93.7\pm1.66$\\
        & \checkmark & $97.04\pm0.98$\\
        \midrule
        \multirow{2}{0.12\textwidth}{$k=80$}  
        & \xmark & $97.38\pm1.48$\\
        & \checkmark & $98.09\pm0.77$\\
        \bottomrule
    \end{tabular}
    \vspace{-0.4cm}
\end{table}

In this experiment, we have also considered larger values of $k$ such as 20, 40, and 80, along with considering the entire test set containing 3629 images. The trained models have been tested using all the samples from the meta-testing set. After domain adaptation, the models were already capable of distinguishing different types of leaf disease. Hence, even though it never saw any tomato leaf samples, it could achieve high accuracy like 89.06\%, 92.46\%, 94.07\% respectively, for the 5-shot, 10-shot, and 15-shot classification tasks. The results were even more promising as the number of samples increased. We achieved 98.09\% accuracy with $20\times$ fewer data compared to the existing works. One obvious observation was that, as the value of $k$ goes higher, the significance of Domain adaptation becomes lower. 
% Finally, to find the maximum capacity of the pipeline, we provided the model with all the $14531$ training samples and evaluated it with all the available samples of the test set ($Q=Full$); which achieved a remarkable accuracy of $99.31\pm0.12$. 
Our conclusion from this experiment is that domain adaptation is an effective approach that enables the model to use $20\times$ fewer data, compromising only 1.22\% accuracy.

\subsection{Performance comparison with Related Works}
\vspace{-1mm}
To evaluate the robustness of our proposed pipeline in different circumstances, we conducted experiments in single-domain and cross-domain environments. In the first experiment, we used the dataset of the few-shot task proposed by Argueso \etal \cite{0018argueso2020fewshot}. In this task, the target classes contained 6 leaf diseases of apple, blueberry, and cherry plants. The experiments were conducted on $Q=50$ for 20 randomly sampled tasks. We applied the model under similar experimental setups, and it significantly outperformed the state-of-the-art. The results are mentioned in \tableautorefname~\ref{tab:arguesoComparison}.

\begin{table}[htb]
    \centering
    \caption{Performance Comparison with Related works on PlantVillage Dataset}
    \vspace{-2mm}
    \label{tab:arguesoComparison}
    \begin{tabular}{L{0.25\textwidth} 
                    C{0.12\textwidth}
                    C{0.12\textwidth} C{0.12\textwidth} C{0.12\textwidth}}

        % \begin{tabular}{L{2.3cm} C{1cm}
                    % C{1cm} C{1cm} C{1cm}}
            \toprule
            \textbf{Approach} 
            % & \boldmath$k=1$ & \boldmath$k=10$ 
            % & \boldmath$k=15$ & \boldmath$k=80$\\
            & \textbf{\textit{k} = 1} & \textbf{\textit{k} = 10} 
            & \textbf{\textit{k} = 15} & \textbf{\textit{k} = 80}\\
            \midrule
            Argüeso \etal \cite{0018argueso2020fewshot}
            & 55.5 & 77.0 & 80.0 & 90.0\\
            Wang \etal \cite{0510wang2021IMAL}
            & 63.8 & \xmark & 91.3 & 96.0\\
            \midrule
            Ours 
            & \boldmath$74.7\pm3.12$ & \boldmath$99.73\pm0.24$
            & \boldmath$99.92\pm0.09$ & \boldmath$99.9\pm0.07$\\
            \midrule
            
    \end{tabular}
    \vspace{-2mm}

\end{table}

\begin{table}[tb]
    \centering
    \caption{Single-Domain and Cross-Domain Experiments}
    \vspace{-2mm}
    \label{tab:singleCrossDomain}
    \begin{tabular}{L{0.1\textwidth} L{0.23\textwidth}         
                    C{0.17\textwidth} C{0.17\textwidth} C{0.17\textwidth}}
    % \begin{tabular}{L{1cm} L{2cm}         
    %                 C{1.2cm} C{1cm} C{1cm}}
        \toprule
        \textbf{$k$-shots} & \textbf{Approach}
        & \textbf{Single-Mixed Domain} & \textbf{Cross-Domain 1}
        & \textbf{Cross-Domain 2}\\
        \midrule
        \multirow{3}{0.15\textwidth}{\textit{K} = 1} & 
        Li \etal \cite{0059yangLi2021metaLearning} & 81.1 & 72.1 & 44.1\\
        & Nuthalapati \etal \cite{0501Nuthalapati2021multi} &
        84.1 & 79.2 & 52 \\
        & Ours & \boldmath$95.56\pm0.91$ & \boldmath$90.29\pm0.91$ & \boldmath$90.05\pm1.06$\\
        \midrule
        \multirow{3}{0.15\textwidth}{\textit{K} = 5} & 
        Li \etal \cite{0059yangLi2021metaLearning} & 87 & 84.9 & 53.1\\
        & Nuthalapati \etal \cite{0501Nuthalapati2021multi} &
        91.2 & 93.7 & 66.5 \\
        & Ours & \boldmath$99.97\pm0.03$ & \boldmath$99.8\pm0.1$ & \boldmath$99.51\pm0.14$\\
        \midrule
        \multirow{3}{0.15\textwidth}{\textit{K} = 10} &       
        Li \etal \cite{0059yangLi2021metaLearning} & 90.4 & 87.1 & 55.8\\
        & Nuthalapati \etal \cite{0501Nuthalapati2021multi} &
        92.9 & 95.5 & 71.6\\
        & Ours & \boldmath$99.97\pm0.03$ & \boldmath$99.89\pm0.05$ & \boldmath$99.41\pm0.28$\\
        \midrule
    \end{tabular}
    % \vspace{-0.4cm}
    
\end{table}

Afterward, the robustness of the pipeline was shown with the single-domain and cross-domain tasks as proposed in \cite{0059yangLi2021metaLearning}. For this task, the `Plants \& Pest' dataset has been utilized. Following the authors \cite{0059yangLi2021metaLearning}, this experiment has been conducted keeping three tasks in mind, named as, single-mixed domain, cross-domain1, and cross-domain2 (results in \tableautorefname~\ref{tab:singleCrossDomain}). In the single-mixed domain experiment, the source class contained samples of both leaves and pests. Although the target classes contained different types of leaves or pests, the classification task was simpler. In the cross-domain 1 experiment, the pest classes were chosen as the source class and the plant classes as the target class. The cross-domain 2 experiment was the other way around. The merit of these experiments is that, if the model can learn leaves after only being trained on pest samples, it signifies that it has learned very good feature representation. Regardless of the single or cross-domain tasks, the pipeline has outperformed the works of \cite{0059yangLi2021metaLearning,0510wang2021IMAL} by a good margin. This shows the generalizing ability of the model under different circumstances. %with just a few shots.

% \subsection{Error Analysis}

The existing literature %related works that we compared 
used samples that were mostly taken in laboratory conditions. Hence, to further investigate the robustness of the pipeline in different field conditions, we tested the performance on two more datasets, namely the `Potato Leaf Disease’ \cite{potato5cls} and `Cotton Leaf Disease’ \cite{cotton4cls}. To the best of our knowledge, this is the first few-shot benchmark for these two datasets. We took the results for $k=1,5,10, \& 15$, considered all the remaining images as Query samples, and repeated the evaluation for 100 tasks. The findings presented in \tableautorefname~\ref{tab:fieldImg} shows that the proposed pipeline has achieved satisfactory performance even with the samples taken under real-world field conditions with diversified backgrounds. 
\vspace{-2mm}

\begin{table}[tb]
    \centering
    \caption{Performance comparison on Field Conditions}
    \vspace{-2mm}
    \label{tab:fieldImg}
    % \begin{tabular}{L{0.1\textwidth} L{0.23\textwidth}         
    %                 C{0.17\textwidth} C{0.17\textwidth} C{0.17\textwidth}}
    \begin{tabular}{L{1.3cm} C{3cm}         
                    C{3cm}}
        \toprule
        \textbf{$k$-shot} & \textbf{Potato Disease Dataset}
        & \textbf{Cotton Disease Dataset}\\
         \midrule
        $k=1$ & $48.61\pm3.68$ & $63.88\pm2.54$\\
        $k=5$ & $76.96\pm1.84$   & $97.58\pm0.45$\\
        $k=10$ & $89.61\pm2.03$ & $99.40\pm0.49$\\
        $k=15$ & $92.95\pm2.07$ & $99.28\pm0.56$\\
        
        % \midrule

        \midrule
    \end{tabular}
        \vspace{-0.5cm}
\end{table}

%----------------------------------------------------------------------
\section{Conclusion}
% \vspace{-1mm}
The limited availability of large-scale, diverse leaf disease datasets has motivated the development of models capable of achieving high accuracy with a minimal amount of samples. In this work, we proposed \archiName, an FSL pipeline for leaf disease classification under data-scarce conditions. Our approach leverages a CNN-based feature extraction block composed of nine state-of-the-art architectures, each contributing a distinct representation of the input. These multi-perspective features are fused and processed by a Bi-LSTM-based classifier for final prediction. To enhance generalization to unseen classes, we incorporate domain adaptation, enabling feature extractors to capture domain-specific cues. The proposed method achieves competitive accuracy substantially outperforming baseline FSL approaches. The strength of \archiName\ is further demonstrated through solid performance on mixed-domain and cross-domain tasks. This work offers a scalable and data-efficient solution for plant disease diagnosis, with potential applications in real-world agricultural settings. Future work could extend the framework to few-shot segmentation or detection and explore lightweight feature extractors for deployment on edge devices.

\bibliographystyle{splncs04}
\bibliography{egbib}

\begin{thebibliography}{10}
\providecommand{\url}[1]{\texttt{#1}}
\providecommand{\urlprefix}{URL }
\providecommand{\doi}[1]{https://doi.org/#1}

\bibitem{abbas2021tomato}
Abbas, A., Jain, S., Gour, M., Vankudothu, S.: {Tomato plant disease detection using transfer learning with C-GAN synthetic images}. Computers and Electronics in Agriculture  \textbf{187},  106279 (2021)

\bibitem{mehedi2024fslBHDR}
Ahamed, M., Kabir, R.B., Dipto, T.T., Al~Mushabbir, M., Ahmed, S., Kabir, M.H.: Performance analysis of few-shot learning approaches for bangla handwritten character and digit recognition. In: 2024 6th International Conference on Sustainable Technologies for Industry 5.0 (STI). pp.~1--6 (2024)

\bibitem{ahmed2022classification}
Ahmed, S.: Classification of Plant Disease from Leaf Images Using Few-Shot Learning. Msc thesis, Department of Computer Science and Engineering (CSE), Islamic University of Technology (2022)

\bibitem{9810234}
Ahmed, S., Hasan, M.B., Ahmed, T., Sony, M.R.K., Kabir, M.H.: Less is more: Lighter and faster deep neural architecture for tomato leaf disease classification. IEEE Access  \textbf{10},  68868--68884 (2022)

\bibitem{ahmed2024exeNet}
Ahmed, T., Hasan, M.B., Ahmed, S., Kabir, M.H.: Exe-net: Explainable ensemble network for potato leaf disease classification. In: 2024 IEEE Canadian Conference on Electrical and Computer Engineering (CCECE). pp. 335--339 (2024)

\bibitem{alanezi2022livestock}
Alanezi, M.A., Shahriar, M.S., Hasan, M.B., Ahmed, S., Sha’aban, Y.A., Bouchekara, H.R.E.H.: Livestock management with unmanned aerial vehicles: A review. IEEE Access  \textbf{10},  45001--45028 (2022)

\bibitem{0018argueso2020fewshot}
Argüeso, D., Picon, A., Irusta, U., Medela, A., San-Emeterio, M.G., Bereciartua, A., Alvarez-Gila, A.: Few-shot learning approach for plant disease classification using images taken in the field. Computers and Electronics in Agriculture  \textbf{175},  105542 (2020)

\bibitem{rahman2022twoDecades}
Ashikur~Rahman, A.B.M., Hasan, M.B., Ahmed, S., Ahmed, T., Ashmafee, M.H., Kabir, M.R., Kabir, M.H.: Two decades of bengali handwritten digit recognition: A survey. IEEE Access  \textbf{10},  92597--92632 (2022)

\bibitem{ashmafee2023apple}
Ashmafee, M.H., Ahmed, T., Ahmed, S., Hasan, M.B., Jahan, M.N., Ashikur~Rahman, A.: An efficient transfer learning-based approach for apple leaf disease classification. In: International Conference on Electrical, Computer and Communication Engineering (ECCE). pp.~1--6 (2023)

\bibitem{buja2021advances}
Buja, I., Sabella, E., Monteduro, A.G., Chiriacò, M.S., De~Bellis, L., Luvisi, A., Maruccio, G.: Advances in plant disease detection and monitoring: From traditional assays to in-field diagnostics. Sensors  \textbf{21}(6) (2021)

\bibitem{0103chen2019aCloseLook}
Chen, W., Liu, Y., Kira, Z., Wang, Y.F., Huang, J.: A closer look at few-shot classification. CoRR  \textbf{abs/1904.04232} (2019)

\bibitem{rafi2025mangoLeafVit}
Chowdhury, R.H., Ahmed, S.: Mangoleafvit: Leveraging lightweight vision transformer with runtime augmentation for efficient mango leaf disease classification. In: 2024 27th International Conference on Computer and Information Technology (ICCIT). pp. 699--704 (2024). \doi{10.1109/ICCIT64611.2024.11022482}

\bibitem{1018dvornik2020selectingRelevant}
Dvornik, N., Schmid, C., Mairal, J.: Selecting relevant features from a multi-domain representation for few-shot classification. In: Computer Vision -- ECCV 2020. pp. 769--786. Springer International Publishing (2020)

\bibitem{0509itziar2022analysis}
Egusquiza, I., Picon, A., Irusta, U., Bereciartua-Perez, A., Eggers, T., Klukas, C., Navarra-Mestre, R.: Analysis of few-shot techniques for fungal plant disease classification and evaluation of clustering capabilities over real datasets. Frontiers in Plant Science  \textbf{13} (2022)

\bibitem{1011chelsea2017model}
Finn, C., Abbeel, P., Levine, S.: Model-agnostic meta-learning for fast adaptation of deep networks. In: Proceedings of the 34th International Conference on Machine Learning. vol.~70, pp. 1126--1135. PMLR (06--11 Aug 2017)

\bibitem{graves2005framewise}
Graves, A., Schmidhuber, J.: Framewise phoneme classification with bidirectional lstm and other neural network architectures. Neural networks  \textbf{18}(5-6),  602--610 (2005)

\bibitem{he2016deep}
He, K., Zhang, X., Ren, S., Sun, J.: {Deep Residual Learning for Image Recognition}. In: IEEE Conference on Computer Vision and Pattern Recognition (CVPR). pp. 770--778 (2016)

\bibitem{herok2023cotton}
Herok, A., Ahmed, S.: Cotton leaf disease identification using transfer learning. In: 2023 International Conference on Information and Communication Technology for Sustainable Development (ICICT4SD). pp. 158--162 (2023)

\bibitem{0017hu2019aLowShot}
Hu, G., Wu, H., Zhang, Y., Wan, M.: A low shot learning method for tea leaf’s disease identification. Computers and Electronics in Agriculture  \textbf{163},  104852 (2019)

\bibitem{huang2017densely}
Huang, G., Liu, Z., Van Der~Maaten, L., Weinberger, K.Q.: {Densely Connected Convolutional Networks}. In: IEEE Conference on Computer Vision and Pattern Recognition (CVPR). pp. 2261--2269 (2017)

\bibitem{hughes2015open}
Hughes, D.P., Salath{\'{e}}, M.: {An open access repository of images on plant health to enable the development of mobile disease diagnostics through machine learning and crowdsourcing}. arXiv - Computing Research Repository  (2015)

\bibitem{1028jamal2019taskAgnostic}
Jamal, M.A., Qi, G.J.: Task agnostic meta-learning for few-shot learning. In: Proceedings of the IEEE/CVF Conference on Computer Vision and Pattern Recognition (CVPR) (2019)

\bibitem{khan2022rethinking}
Khan, A.M., Ashrafee, A., Sayera, R., Ivan, S., Ahmed, S.: Rethinking cooking state recognition with vision transformers. In: 25th International Conference on Computer and Information Technology (ICCIT). pp. 170--175 (2022)

\bibitem{0101koch2015siamese}
Koch, G., Zemel, R., Salakhutdinov, R.: Siamese neural networks for one-shot image recognition. In: Proceedings of the 32nd Int. Conference on Machine Learning. {JMLR} (2015)

\bibitem{1017Kolesnikov2020bigTransfer}
Kolesnikov, A., Beyer, L., Zhai, X., Puigcerver, J., Yung, J., Gelly, S., Houlsby, N.: Big transfer (bit): General visual representation learning. In: ECCV 2020. pp. 491--507 (2020)

\bibitem{li2021plant}
Li, L., Zhang, S., Wang, B.: {Plant Disease Detection and Classification by Deep Learning—A Review}. IEEE Access  \textbf{9},  56683--56698 (2021)

\bibitem{0059yangLi2021metaLearning}
Li, Y., Yang, J.: Meta-learning baselines and database for few-shot classification in agriculture. Computers and Electronics in Agriculture  \textbf{182},  106055 (2021)

\bibitem{maeda2020comparison}
Maeda-Gutierrez, V., Galvan-Tejada, C.E., Zanella-Calzada, L.A., Celaya-Padilla, J.M., Galv{\'a}n-Tejada, J.I., Gamboa-Rosales, H., Luna-Garcia, H., Magallanes-Quintanar, R., Guerrero~Mendez, C.A., Olvera-Olvera, C.A.: {Comparison of Convolutional Neural Network Architectures for Classification of Tomato Plant Diseases}. Applied Sciences  \textbf{10}(4) (2020)

\bibitem{morshed2022Fruit}
Morshed, M.S., Ahmed, S., Ahmed, T., Islam, M.U., Ashikur~Rahman, A.: Fruit quality assessment with densely connected convolutional neural network. In: 2022 12th International Conference on Electrical and Computer Engineering (ICECE). pp.~1--4 (2022)

\bibitem{0070sergey2021image}
Nesteruk, S., Shadrin, D., Pukalchik, M.: Image augmentation for multitask few-shot learning: Agricultural domain use-case. CoRR  \textbf{abs/2102.12295} (2021)

\bibitem{1029nichol2018reptile}
Nichol, A., Achiam, J., Schulman, J.: On first-order meta-learning algorithms. CoRR  \textbf{abs/1803.02999} (2018)

\bibitem{cotton4cls}
Noon, S.K., Amjad, M., Ali~Qureshi, M., Mannan, A.: Computationally light deep learning framework to recognize cotton leaf diseases. Journal of Intelligent \& Fuzzy Systems  \textbf{40},  12383--12398 (2021), 6

\bibitem{0501Nuthalapati2021multi}
Nuthalapati, S.V., Tunga, A.: Multi-domain few-shot learning and dataset for agricultural applications. In: Proceedings of the IEEE/CVF International Conference on Computer Vision (ICCV) Workshops. vol. abs/2109.09952, pp. 1399--1408 (October 2021)

\bibitem{panno2021review}
Panno, S., Davino, S., Caruso, A.G., Bertacca, S., Crnogorac, A., Mandić, A., Noris, E., Matić, S.: {A Review of the Most Common and Economically Important Diseases That Undermine the Cultivation of Tomato Crop in the Mediterranean Basin}. Agronomy  (2021)

\bibitem{rafi2023pest}
Rafi, M.H., Ratul~Mahjabin, M., Rahman, M.S., Hasanul~Kabir, M., Ahmed, S.: A critical analysis of deep learning applications in crop pest classification: Promising pathways and limitations. In: 2023 26th International Conference on Computer and Information Technology (ICCIT). pp.~1--6 (2023)

\bibitem{raiyan2025hasper}
Raiyan, S.R., Amio, Z.Z., Ahmed, S.: Hasper: An image repository for hand shadow puppet recognition. In: ICCV 2025 Workshop on Cultural Continuity of Artists (2025)

\bibitem{potato5cls}
Sholihati, R.A., Sulistijono, I.A., Risnumawan, A., Kusumawati, E.: Potato leaf disease classification using deep learning approach. In: International Electronics Symposium (IES). pp. 392--397 (2020)

\bibitem{simonyan2014very}
Simonyan, K., Zisserman, A.: Very deep convolutional networks for large-scale image recognition. arXiv preprint arXiv:1409.1556  (2014)

\bibitem{1004snell2017prototypical}
Snell, J., Swersky, K., Zemel, R.: Prototypical networks for few-shot learning. In: Advances in Neural Information Processing Systems. vol.~30. Curran Associates, Inc. (2017)

\bibitem{sun2024fslLeafSurvey}
Sun, J., Cao, W., Fu, X., Ochi, S., Yamanaka, T.: Few-shot learning for plant disease recognition: A review. Agronomy Journal  \textbf{116}(3),  1204--1216 (2024)

\bibitem{7780677}
Szegedy, C., Vanhoucke, V., Ioffe, S., Shlens, J., Wojna, Z.: Rethinking the inception architecture for computer vision. In: IEEE Conference on Computer Vision and Pattern Recognition (CVPR). pp. 2818--2826 (2016)

\bibitem{0104yonglong2020rethingkingFewShot}
Tian, Y., Wang, Y., Krishnan, D., Tenenbaum, J.B., Isola, P.: Rethinking few-shot image classification: A good embedding is all you need? In: Computer Vision -- ECCV 2020. pp. 266--282. Springer International Publishing (2020)

\bibitem{UskanerHepsağ2024}
Uskaner~Hepsa{\u{g}}, P.: Efficient plant disease identification using few-shot learning: a transfer learning approach. Multimedia Tools and Applications  \textbf{83}(20),  58293--58308 (Jun 2024)

\bibitem{1012vinyals2016matching}
Vinyals, O., Blundell, C., Lillicrap, T., kavukcuoglu, k., Wierstra, D.: Matching networks for one shot learning. In: Advances in Neural Information Processing Systems. vol.~29. Curran Associates, Inc. (2016)

\bibitem{0050wang2019plant}
Wang, B., Wang, D.: Plant leaves classification: A few-shot learning method based on siamese network. IEEE Access  \textbf{7},  151754--151763 (2019)

\bibitem{0510wang2021IMAL}
Wang, Y., Wang, S.: Imal: An improved meta-learning approach for few-shot classification of plant diseases. In: IEEE 21st International Conference on Bioinformatics and Bioengineering (BIBE). pp.~1--7 (2021)

\bibitem{xu2023}
Xu, M., Kim, H., Yang, J., Fuentes, A., Meng, Y., Yoon, S., Kim, T., Park, D.S.: Embracing limited and imperfect training datasets: opportunities and challenges in plant disease recognition using deep learning. Frontiers in Plant Science  \textbf{Volume 14 - 2023} (2023)

\bibitem{yasmeen2021csvcNet}
Yasmeen, A., Rahman, F.I., Ahmed, S., Kabir, M.H.: Csvc-net: Code-switched voice command classification using deep cnn-lstm network. In: 2021 Joint 10th International Conference on Informatics, Electronics \& Vision (ICIEV) and 2021 5th International Conference on Imaging, Vision \& Pattern Recognition (icIVPR). pp.~1--8 (2021)

\bibitem{zabihzadeh2024zs}
Zabihzadeh, D., Masoudifar, M.: Zs-dml: Zero-shot deep metric learning approach for plant leaf disease classification. Multimedia Tools and Applications  \textbf{83}(18),  54147--54164 (2024)

\bibitem{zhao2025fsl}
Zhao, J., Kong, L., Lv, J.: An overview of deep neural networks for few-shot learning. Big Data Mining and Analytics  \textbf{8}(1),  145--188 (2025)

\end{thebibliography}

\end{document}